\documentclass{article}
\usepackage[utf8]{inputenc}
\usepackage{graphicx}

\title{Natural scene statistics mediate the perception of image complexity}
\author{Nicolas Gauvrit, Fernando Soler-Toscano, and Hector
Zenil}
\date{July 28th, 2014}

\begin{document}

\maketitle

\section*{\hfil Abstract \hfil}

Humans are sensitive to complexity and regularity in patterns~\cite{yamada2013pattern, falk1997making}. The subjective perception of pattern complexity is correlated to algorithmic (Kolmogorov-Chaitin) complexity as defined in computer science~\cite{li}, but also to the frequency of naturally occurring patterns~\cite{hsu2010subjective}. However, the possible mediational role of natural frequencies in the perception of algorithmic complexity remains unclear. Here we reanalyze~\cite{hsu2010subjective} through a mediational analysis, and complement their results in a new experiment. We conclude that human perception of complexity seems partly shaped by natural scenes statistics, thereby establishing a link between the perception of complexity and the effect of natural scene statistics.

\smallskip
\noindent \begin{center}\textbf{Keywords:} visual complexity; visual perception;
algorithmic complexity; randomness \end{center}

\section{Introduction}

Humans are extremely sensitive to patterns and regularities~\cite{yamada2013pattern}. Our brains detect slight departures from randomness. Someone throwing 3 dice and getting three ‘6s’ or the pattern ‘1, 2, 3’ is likely to be stunned by the fact that these combinations are regular, to be incredulous in the face of this meager evidence that the dice are fai~\cite{falk1997making}. This general human feature—the discernment of rules governing the world—may be thought of as the cognitive basis of science, but also as an adaptive ability shaped by natural evolution to avoid predictable dangers.
Psychologists have linked our natural perception of randomness to the mathematical theory of algorithmic complexity (also known as Kolmogorov-Chaitin complexity): the more complex the stimulus, the more random it will be perceived to be. Formally, the algorithmic complexity of a sequence is the length of the shortest program that produces the sequence in question and halts~\cite{li}. In this definition, the said program doesn’t involve a specific computer, but rather a general Universal Turing Machine, an abstract computer. Algorithmic complexity is related to the probability that such a machine, fed with a random program, will produce a particular sequence and halt, a link formally proven by the coding theorem~\cite{levin} and initially conceived to solve the problem of induction—which it does in a very general and powerful way~\cite{solomonoff}—so powerful that the measure is indeed ultimately uncomputable, though approximations are possible. The main intuition behind algorithmic probability is that if a sequence is not random then it will contain some regularity that can be encoded in a computer program of length shorter than the sequence that can generate it by mechanistic means. And shorter programs are more likely to occur, and therefore more frequent than longer ones if each program
instruction is uniformly randomly chosen, which establishes a powerful connection between complexity and frequency. Within the framework of algorithmic complexity theory, “randomness” and “complexity” are interchangeable concepts: the formal definition of randomness relies on complexity, and complexity is a direct measure of randomness.

The hypothesis that the human perception of randomness is linked to algorithmic complexity could not be verified prior to recent developments in computer science. Indeed, if methods have long existed that allow satisfactory estimations of the algorithmic randomness of long sequences, such as compression algorithms~\cite{ziv1978compression}, until recently no such methods were available to assess the algorithmic complexity of short sequences~\cite{soler2013correspondence, soler2014calculating, zenil2012two, gauvrit2014algorithmic}.

In light of algorithmic complexity, we undertook an investigation into how humans perceive randomness, and how we learn (if we do) to perceive complexity. Hsu, Griffiths and Schreiber~\cite{hsu2010subjective} advanced an interesting hypothesis: the frequency with which a pattern appears in real world scenes could explain how we perceive randomness, permitting us to infer complexity from the world we see. Hsu et al.~\cite{hsu2010subjective} scanned a set of photographs of real world natural scenes and extracted every possible $4 \times 4$ array from these images. Then they computed the resulting probability distribution, and derived the randomness of each array $x$, defined as $random(x) = log(P(x|r)/P(x|n)), P(x|n)$ being the relative frequency of the array in the natural scene database, and $P(x|r)$ the probability that this array appears by chance if every cell in the array is selected at random (either white or black). They chose 100 balanced arrays with probabilities of occurrence in real scenes ranging from low to high. They then had 77 subjects decide whether these arrays looked random or not. This led to a measure of subjective randomness (the proportion of participants declaring the array random) for each array.
They found that subjective probability and natural randomness were positively correlated on these
particular 100 arrays $(r = .75, p < .0001)$. We computed that Kolmogorov-Chaitin complexity is
also significantly linked to the subjective perception of randomness. With the arrays published in Hsu et
al.~\cite{hsu2010subjective}, we found a correlation of $r = .52 (p < .0001)$ between two-dimensional algorithmic complexity as defined in Zenil et al.~\cite{zenil2012two} —see also Gauvrit, Zenil, Delahaye and Soler-Toscano~\cite{soler2013correspondence}—and subjective probability. This pattern of correlations is not surprising. Because the world can be thought of as a generator of patterns, like a random computer program, the probability that an array will occur in the world is then linked to its algorithmic complexity. Indeed, as computed with the 100 arrays of Hsu et al.~\cite{hsu2010subjective}, we also found a positive correlation between algorithmic complexity and natural scene statistics $(r = .50, p < .0001)$.
Could natural scene statistics account for human perception of algorithmic complexity? This would be in line with recent results in neuroscience reported by Berkes, Orban, Lengyel and Friser~\cite{berkes2011spontaneous}. They analyzed cortical activity in ferrets, and compiled evidence in favor of the hypothesis that our brain learns an optimal internal probabilistic model of the environment, based on natural world frequencies—see also~\cite{teglas2011pure} for examples of children’s rapid adaptation to natural frequencies. But how much of our perception of randomness is attributable to learning through the natural world? To answer this question, we performed a mediation analysis using scaled data. A regression of subjective randomness on both algorithmic complexity and natural scenes statistics gives an adjusted R-squared equal to $.58 (p < .0001)$. Figure 1(A) displays the coefficients linking complexity to subjective randomness $(.19, p = .013)$ and natural scenes statistics to subjective randomness, controlling for algorithmic complexity $(.66, p < .0001)$. In this figure, “Algorithmic complexity” stands for the Kolmogorov-Chaitin complexity of the arrays, as approximated by the method described in Zenil et al. \cite{zenil2012two}. “Natural statistics” refers to the random function defined above, in which $P(x|n)$ stands for the frequency of array $x$ in the natural scenes dataset. Last, “subjective randomness” is a shorthand for $log(p(x))$, where $p(x)$ designates the proportion of participants who indicate that $x$ is seemingly random. A Sobel test confirms the mediational role of natural scene statistics $(z = 4.78, p < .0001)$.

\begin{figure}
\centering
\includegraphics[scale=0.4]{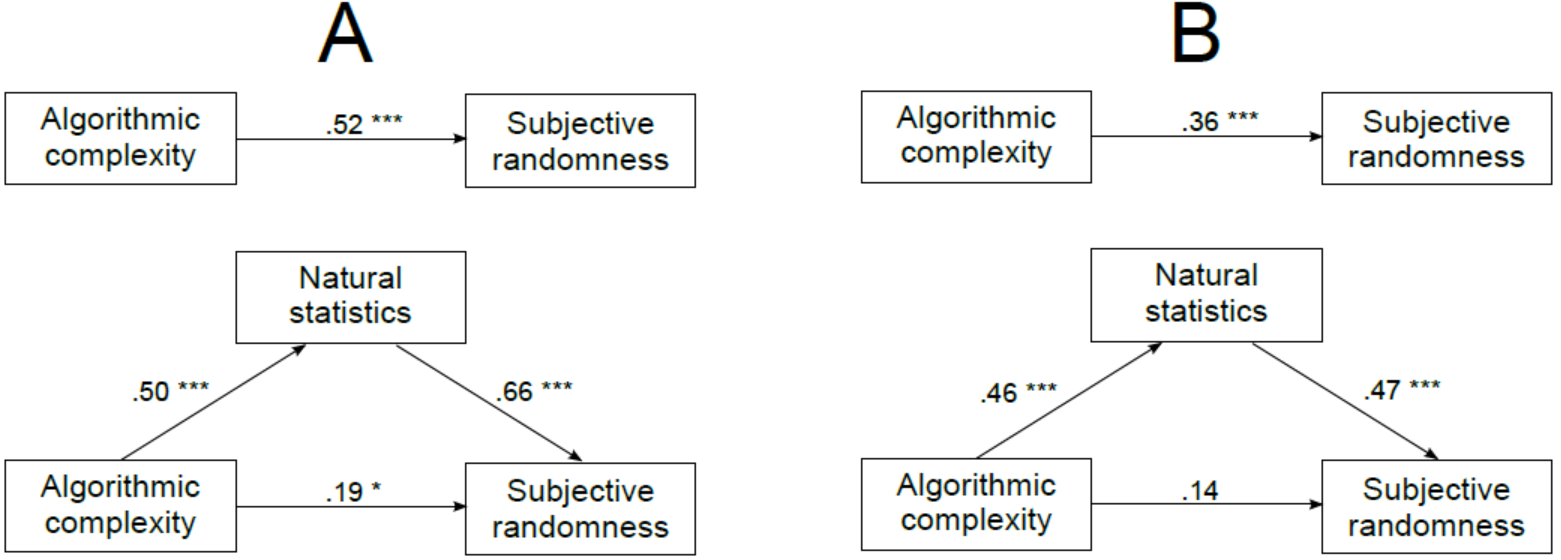}
\caption{Mediation analysis, performed with scaled data as computed from the dataset of Hsu et al. (2010) [subplot A] and with our experimental data [subplot B]. In each subplot, the top graph displays the standardized correlation coefficient. The bottom graph displays (1) the standardized regression coefficient between natural statistics and algorithmic complexity, (2) the standardized regression coefficient between complexity and subjective randomness, and (3) the partial standardized regression
coefficient between natural scenes statistics and subjective randomness, controlling for algorithmic
complexity. * $p < .05$ , *** $p < .001$.}
\label{fig:mediation-analysis}
\end{figure}

The result suggests that our perception of complexity is partially driven by the perception of natural scenes. However, it is fair to underscore two points that may prejudice the values found here. First, we cannot control the link between “natural scene statistics” (i.e. the “random” function) and the choice of the set of pictures. Second, because the 100 arrays chosen for use here fall within certain parameters (they are all balanced, and have been chosen in such a way that they are evenly distributed on the natural scene statistics scale), variance of natural scene statistics could be artificially high. In the following experiment, we overcome these two possible drawbacks in order to get a clear view of the possible mediational role of natural scene statistics in the perception of complexity.

\section{Method}
We perform an experiment similar to the one presented above, but releasing some constraints that could affect the results. We do not impose that every pattern is balanced in terms of white and black cells. We do not choose still nature shots only. Our hypothesis is that even when these constraints are relieved, natural scene statistics will play a mediational role.

\subsection{Participants}
A sample of 100 participants (59 male, 41 female) was recruited via the Amazon Mechanical Turk. Hired “workers” from the Mechanical Turk were required to have a 90\% approval rating on previous Mechanical Turk tasks (HITs) and at least 50 previous HITs approved. Ages in years ranged between 19 and 55 $(mean \pm sd = 30.6 \pm 8)$. Participants were paid 0.30 USD for their participation. The experiment duration ranged from 84s to 289s $(mean \pm sd = 212.2 \pm 45)$. Older participants in this sample showed a slight tendency to need more time $(r = .15)$.

\subsection{Stimuli}

Hsu et al.~\cite{hsu2010subjective} used a set of 62 pictures previously used by Doi, Inui, Lee, Wachtler and Sejnowski~\cite{inui2003spatiochromatic} to compute the natural scenes statistics. All pictures were still nature shots, including no faces, urban scenes or artificial objects. Therefore, the random function may vary if computed with other sets of pictures. To test this hypothesis, we applied the method used by Hsu et al.~\cite{hsu2010subjective} to a new set of 100 random pictures, taken from the Wikimedia Commons database\footnote{http://commons.wikimedia.org/wiki/Special:Random/Image}. The sample included natural scenes but also animals and non-natural objects such as buildings. Then we binarized the pictures to black and white pixels using the median as the threshold. We then divided each image into $4 \times 4$ adjacent binary square arrays and calculated (for the whole set of 100 images) the probability of each square.
The resulting “random” function is strongly correlated to the data obtained by Hsu et al.~\cite{hsu2010subjective}when computed on their choice of 100 arrays $(r = .91, p < .0001)$, which validates the method. The correlation between natural scenes statistics (function random) and algorithmic complexity was .50 when computed on the 100 arrays chosen by Hsu et al~\cite{hsu2010subjective}. However, because the choice of arrays was not random, the correlation could well be overestimated. When computed on every possible $4 \times 4$ array found while scanning the 100 random pictures, the correlation remains highly significant, although slightly lower $(r = .42, p < .0001)$, confirming the previous result.
We then picked at random a sample of 100 arrays from among all the arrays found in our set of 100 images, using the sample function in R. We did not contrive to obtain balanced arrays, a departure from the design of Hsu et al.~\cite{hsu2010subjective}. Figure 2 displays the 100 arrays obtained by random selection.

\begin{figure}
\centering
\includegraphics[scale=0.25]{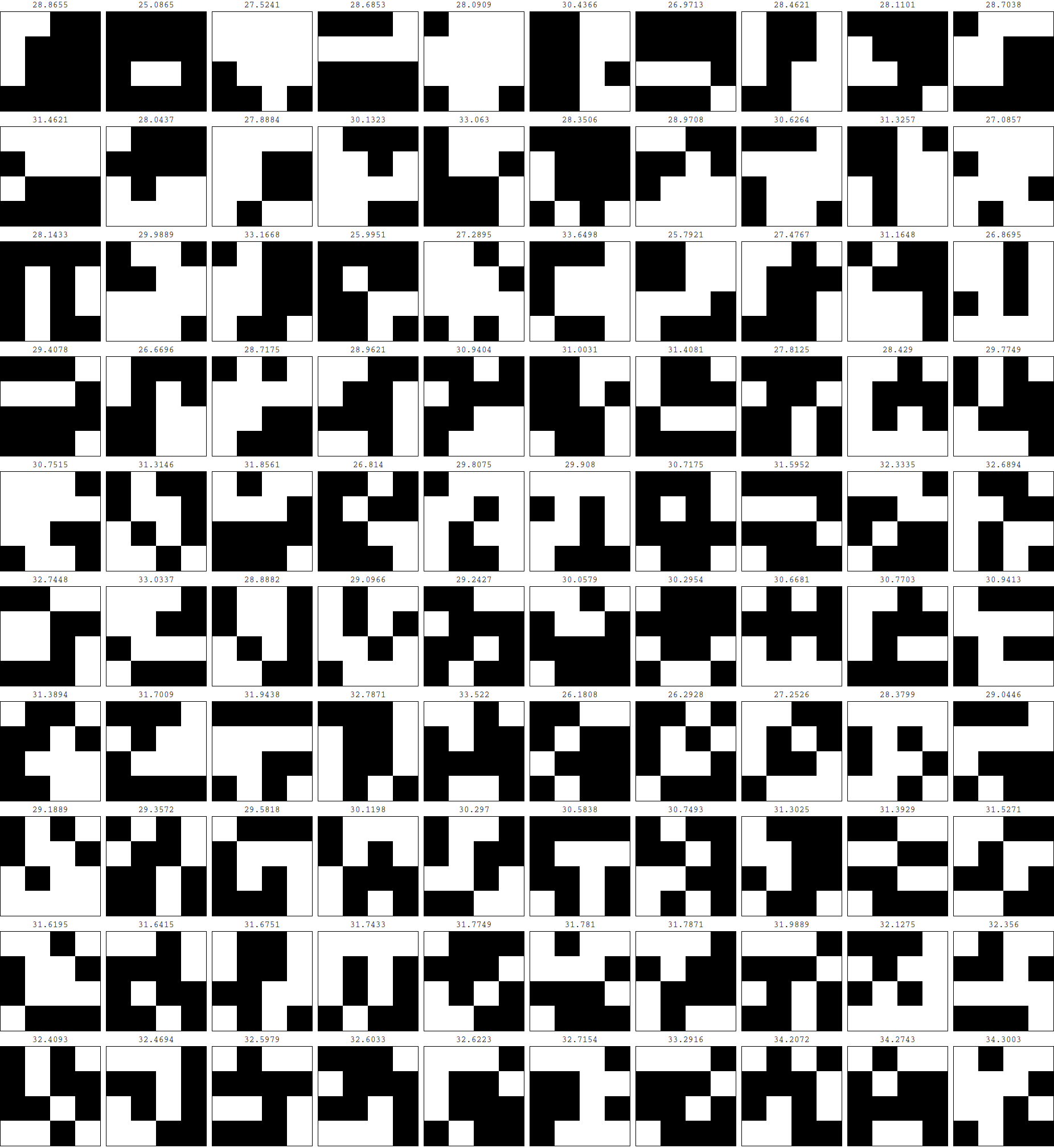}
\caption{The 100 arrays used in our experiment together with their algorithmic complexity (above each array). Arrays are ordered according to their natural scene frequency (from more to less frequent arrays).}
\label{fig:my_label}
\end{figure}

\subsection{Procedure}
The procedure mirrored the one used in Hsu et al. (2010), although our experiment took place online. Participants filled out a questionnaire similar to that used by Hsu et al. (2010). They were informed that a series of arrays would appear on the screen, and that their task was to decide whether the arrays were produced by a random process or by a nonrandom process. For each array, they were asked to press a button, either “random” or “not random” according to their perception.

\subsection{Results}

The data were analyzed with the same method as Hsu et al.~\cite{hsu2010subjective}. Algorithmic complexity is positively correlated with natural statistics $(r = .46, p < .0001)$ and subjective randomness $(r = .36, p < .0001)$, as are subjective randomness and natural statistics $(r = .56, p < .0001)$.
A multiple regression of subjective randomness on algorithmic complexity and natural scene statistics yields an adjusted R-squared of $.31 (p < .0001)$. Figure 1(B) displays the coefficients linking complexity to subjective randomness $(.14, p = .15)$ and natural scenes statistics to subjective randomness, controlling for algorithmic complexity $(.47, p < .0001)$. A Sobel test confirms the mediational role of natural scene statistics $(z = 3.66, p < .001)$.

\newpage

\section{Discussion}

Perhaps as an upshot of the eschewal of constraints as compared with the Hsu et al.~\cite{hsu2010subjective} study (balanced arrays scattered along the natural probability range), the coefficients are now smaller. However, the patterns of correlations are remarkably similar.
These results suggest that natural scene statistics are indeed an important element in the perception of complexity. The correspondence between the reanalysis and the subsequent experiment also suggests that there is some objective natural probability of arrays, linked both to our perception of complexity and to the formal definition of complexity arising from Kolmogorov-Chaitin theory. Although our perception of complexity may be largely explained by natural scene statistics, this does not preempt the possibility of a complementary means of perception, which could eventually turn out to be innate. However, further studies would be
needed to confirm this assumption.


\bibliographystyle{alpha}
\bibliography{bibliography}

\end{document}